\documentclass[10pt,twocolumn,letterpaper]{article}
\usepackage[pagenumbers]{cvpr}
\usepackage{cuted}
\usepackage{currfile}
\usepackage{caption}
\usepackage{booktabs}
\usepackage{multirow}
\usepackage{xspace}
\usepackage{amsmath}

\newcommand{\qwen}{Qwen3-VL\xspace}
\newcommand{\qwenfull}{Qwen3-VL-30B-A3B\xspace}
\newcommand{\kimi}{Kimi-VL\xspace}
\newcommand{\kimifull}{Kimi-VL-A3B\xspace}
\newcommand{\gemma}{Gemma-4\xspace}
\newcommand{\gemmafull}{Gemma-4-26B-A4B\xspace}
\newcommand{\vex}{vision-exclusive\xspace}

\newcommand{\vstar}{V$^{*}$\xspace}

\newsavebox{\fittablebox}
\newcommand{\fittable}[1]{\sbox{\fittablebox}{#1}%
  \typeout{FITTABLE natural=\the\wd\fittablebox\space line=\the\linewidth}%
  \ifdim\wd\fittablebox>\linewidth \resizebox{\linewidth}{!}{\usebox{\fittablebox}}\else\usebox{\fittablebox}\fi}
\newcommand{\coderepo}{\url{https://github.com/gezhixuan/moedep}}

\definecolor{cvprblue}{rgb}{0.21,0.49,0.74}
\usepackage[pagebackref,breaklinks,colorlinks,allcolors=cvprblue]{hyperref}
\def\confName{CVPR}
\def\confYear{2027}

\title{Lit Up but Not Needed: Scoped Lesions Show Where Multimodal Mixture-of-Experts Depend on Their Experts}

\author{Zhixuan Ge \qquad Anqi Li \qquad Kexin Geng \qquad Wei Qiu\textsuperscript{\dag}\\
Rice University\\
{\tt\small \{zg33, al318, kg111, wq8\}@rice.edu} \quad {\small \textsuperscript{\dag}Corresponding author}}

\begin{document}
\maketitle
\begin{abstract}
Cognitive neuroscience learned that an activation map is not a map of necessity. Regions that light up during a task are not always needed for it, because other regions can perform the same function, and only lesions reveal which ones are. Mixture-of-experts (MoE) vision-language models come with an activation map for free, since the router records which experts respond to image and text tokens, and recent work reads modality-specialized experts off this map. We bring the lesion. In \qwenfull, \kimifull and \gemmafull, 3 to 12\% of the experts are vision-exclusive: image tokens route to them at up to five times the uniform rate and text tokens almost never. In \qwen they receive a quarter of all image-token routing, and lesioning all of them costs less than one percentage point of accuracy on average over eight perception benchmarks and free-form OCR, no more than lesioning a random set of the same size. In \kimi and \gemma the lesion also costs no more than a random set. Scoped lesions, which remove an expert set only at image positions or only at text positions, then show where the models do depend on their experts. In \qwen, silencing every expert at every image position costs 3 points, against 45 at text positions, where the dependence sits in dual-use experts that text-only tasks need as well. Across models the image-side dependence is 2 to 3 points in \qwen and \kimi and 26 in \gemma, where the \vex experts are dispensable all the same. Reading text in images is the one ability that needs image-position experts in every model. The exclusive experts receive informative patches, produce outputs of ordinary size, and stay inert under sensory deprivation: text-only training recruits them only when load balancing forces it, with no cost to vision and no gain for text. In \kimi the balancing bias itself manufactures most of them. Load balancing draws the modality map in the part of the computation the models depend on least. Expert selection, pruning and skipping should follow the lesion, not the map.
\end{abstract}

\section{Introduction}
\label{sec:intro}

\begin{figure*}[t]
\centering
\includegraphics[width=\textwidth]{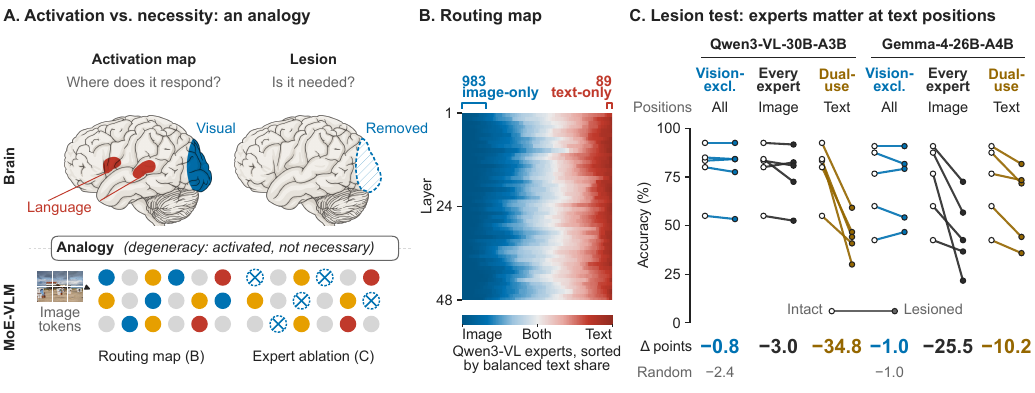}
\caption{\textbf{Experts that image tokens prefer are not the experts that vision needs.}
(A) Cognitive neuroscience separates an activation map (where does a region respond?) from a lesion (is it needed?), and degeneracy makes the two disagree. A MoE router gives the activation map for free, and expert ablation is the lesion. Brain adapted from NIH BioArt \cite{bioart}.
(B) Routing map of \qwen: all 48 layers by 128 experts, sorted within each layer by the fraction of text tokens among the selections of each expert after balancing the token mix. The 983 experts at the left are used almost only by image tokens, 89 at the right almost only by text.
(C) Lesion test. Each line is one perception probe, intact to lesioned. Removing every \vex expert costs at most 1 point on average, no more than a random set of equal size (2.4 and 1.0). Silencing every expert at every image position costs 3.0 points in \qwen and 25.5 in \gemma. Removing the dual-use experts at text positions costs 34.8 and 10.2. \kimi (omitted for space) gives $+1.8$, $-2.0$ and $-1.3$ points.}
\label{fig:teaser}
\end{figure*}

A cortical region that lights up during a task is not always a region the task depends on. Whether the same holds for the experts of a multimodal large language model is the question of this paper. The models we study run on mixture-of-experts (MoE) decoders, in which a router scores each token against a pool of expert feed-forward networks, sends it to the few with the highest scores, and sums their outputs. Because the router chooses for every token at every layer, which experts image tokens and text tokens visit is observable in the forward pass. Hundreds of experts turn out to be visited almost only by image tokens, and recent work reads them off the routing statistics as vision experts that guide expert fine-tuning and routing \cite{expertlens,routingdistraction,smoes}.

In cognitive neuroscience, imaging localizes activity and lesions localize necessity, and the two disagree when a system is degenerate, with structurally different elements able to perform the same function \cite{edelman,pricefriston,rorden}. We ask the lesion's question of multimodal MoEs: where does visual behavior depend on expert computation, and how does that dependence relate to the map the router draws? The answer is a dissociation (\cref{fig:teaser}): the modality map is drawn at image positions, and that is where the models depend on their experts least.

We start from the map. In three production vision-language MoEs, \qwenfull \cite{qwen3vl}, \kimifull \cite{kimivl} and \gemmafull \cite{gemma4}, 3 to 12\% of the routed experts are vision-exclusive: image tokens visit them at up to five times the uniform rate and text tokens almost never, and in \qwen they receive a quarter of all image-token routing. We lesion the whole population and measure what the models lose. On eight perception benchmarks and free-form OCR the lesion leaves accuracy within one percentage point of the intact model on average, and caption likelihood within 0.02 nats per token, no more than a random expert set of the same size. The map is sharp, and the computation behind its vision-exclusive experts is degenerate.

We then locate the expert computation that visual behavior does depend on. A multimodal decoder runs the same experts at image positions, at the question and at the answer it generates, and a lesion removes an expert everywhere it is used, so the three roles are conflated. Neuroscience resolves the same ambiguity by mapping a deficit to the location of the lesion that produced it \cite{bates,rorden}. We introduce its counterpart for a MoE, the scoped lesion, which removes an expert set only at image-token positions or only at text positions, so that a deficit can be attributed to where the experts were used. In \qwen, the experts that text shares with the image top-16 ranking cost 35 points on five perception benchmarks when removed at text positions and 2 points when removed at image positions, and silencing every routed expert at every image position costs 3 points, against 45 for the text positions. The dependence sits with the text, in experts that text-only tasks need as well. Across models, image-position dependence is 2 to 3 points in \qwen and \kimi and 26 in \gemma, where the \vex experts are dispensable all the same. Reading text in images is the one ability that needs image-position expert computation everywhere (\cref{fig:lesion}).

The vision-exclusive experts do real work. They receive the informative patches of an image, respond more to detailed regions, and produce outputs of ordinary size, and their dispensability is degeneracy: other experts can do what they do (\cref{fig:what}). We then ask how the population arises, with the deprivation experiment of cross-modal plasticity \cite{sadato,bavelier,wieselhubel}. Under text-only continued training the vision-exclusive experts are recruited by text only when load balancing forces the router to use them, and the recruitment leaves text quality and vision where they were (\cref{fig:deprivation}). In \kimi, removing the learned balancing bias cuts the population from 144 experts to 53 and changes no behavior (\cref{fig:bias}). Load balancing draws the modality map, by spreading image tokens over experts whose individual contributions any other expert can replace.

This paper makes three contributions: (i) the scoped lesion, which locates where the behavior of a MoE depends on its experts, (ii) a dissociation in three production models, where the \vex experts can be removed at almost no cost and the dependence on experts is largest at text positions, and (iii) the origin of the \vex experts in load balancing. Code is available at \coderepo.

\section{Related Work}
\label{sec:related}

\paragraph{Modality experts in MoE vision-language models.}
MoE decoders are now standard in open vision-language models \cite{qwen3vl,kimivl,gemma4,deepseekvl2,moellava}. Several architectures assign experts to modalities by design \cite{moiie,smoes}. Analyses of models without such constraints report that modality specialization emerges anyway \cite{limoe}. ExpertLens \cite{expertlens} finds a bimodal distribution of image versus text affinity in \qwen and \gemma and selects experts for domain adaptation from the router weights. Routing Distraction \cite{routingdistraction} identifies visual experts by routing frequency and attributes reasoning failures on images to routing away from domain experts. Our work takes the same routing statistics as its starting point and asks what the identified experts are needed for.

\paragraph{Expert specialization in language MoEs.}
Whether routing reflects semantics is contested. Semantic routing is detectable in large MoEs \cite{olson}, but a stable committee of experts serves every domain \cite{committee}, expert usage follows hidden-state geometry rather than domain \cite{myth}, and expert subspaces overlap while still being jointly useful \cite{coherent}.

\paragraph{Circuits across modalities and neuroscience-inspired analysis.}
Vision and text versions of the same task use largely separate circuits in dense VLMs \cite{sametask}, while intermediate representations are shared across modalities \cite{semantichub}. Localizer and lesion methods from cognitive neuroscience have been ported to language models to find language-selective units \cite{alkhamissi} and causally necessary neuron sets per reasoning domain \cite{modular}. Layer perturbations of language models have been mapped onto the lesions of stroke patients \cite{strokelesion}, and critical-period and deprivation experiments have been reproduced in deep networks \cite{kleinman}. We bring the lesion and the deprivation designs to the expert level of multimodal MoEs.

\paragraph{Expert pruning and skipping.}
Post-training expert pruning and skipping select experts by usage or by modality-aware importance \cite{notallexperts,modes}. Our scoped lesions are the case such rules have to handle.

\section{Setup}
\label{sec:setup}

\paragraph{Models.}
\qwenfull \cite{qwen3vl} has 48 MoE layers with 128 routed experts and top-8 softmax routing, trained with a Switch-style auxiliary balancing loss \cite{switch}. \kimifull \cite{kimivl} has 26 MoE layers (the first layer is dense) with 64 routed experts, top-6 sigmoid routing, two always-on shared experts, and DeepSeek-style aux-loss-free balancing through a learned per-expert bias that is added to the routing scores at inference \cite{auxfree,deepseekmoe}. \gemmafull \cite{gemma4} has 30 layers, each with a dense MLP as its shared expert and a parallel MoE block of 128 experts with top-8 softmax routing. Images are encoded at up to 1,280, 1,024 and 280 tokens respectively, and all three models are used as released. We call them \qwen, \kimi and \gemma.

\paragraph{Probes.}
The multimodal set holds 600 image-question pairs, 120 each from POPE \cite{pope} (yes/no hallucination), MMVP \cite{mmvp} (CLIP-blind pairs), \vstar \cite{vstar} (small-object attributes), ZoomBench \cite{zoombench} (fine-detail questions on high-resolution images) and TextVQA \cite{textvqa} (scene text). Harder perception probes add 100 to 120 items each from HR-Bench 4K and 8K \cite{hrbench} and MME-RealWorld-Lite \cite{mmerw}. Text-only inputs are 400 Alpaca chats \cite{alpaca}, 300 of the questions without their image, and ARC-Easy \cite{arc} for text accuracy. Format probes take 300 ASCIIEval concepts \cite{asciieval} as ASCII text, shuffled ASCII, rendered ASCII, caption and photo, plus 48 synthetic shapes as raster, SVG code, ASCII and caption. Multiple-choice and yes/no items are scored from the first answer token, TextVQA by whether the most likely first token begins the gold answer and by VQA soft accuracy of a greedy answer, and captions by the teacher-forced negative log-likelihood of the five COCO references \cite{coco} of each POPE image.

\paragraph{Routing map.}
A forward pass records, for every token and every MoE layer, the top-$k$ experts. Each token carries a modality tag (image, text, or vision delimiter). For expert $e$ in layer $l$ we compute $P(e \mid \text{image})$ and $P(e \mid \text{text})$, the fraction of image-token and text-token selections in that layer that go to $e$, its image load and text load. Uniform routing gives each of the $E$ experts of a layer $1/E$. In \qwen, image tokens outnumber text tokens 18 to 1 in the multimodal set, so the raw text share of ExpertLens \cite{expertlens} puts 60\% of all experts below 5\%. We report the balanced text share, $P(e \mid \text{text}) / (P(e \mid \text{image}) + P(e \mid \text{text}))$, which weighs the two modalities equally. The \vex set is $P(e \mid \text{image}) \geq 10\, P(e \mid \text{text})$ and $P(e \mid \text{image}) \geq 1/E$. The text-exclusive set is its mirror. The image top-16 set contains, per layer, the 16 experts with the highest $P(e \mid \text{image})$ (the 8 highest in \kimi), and splits into a \vex part and a dual-use part. The text top-16 set and the above-uniform set (at least $1/E$ for both modalities) are defined the same way.

\begin{figure*}[t]
\centering
\includegraphics[width=\textwidth]{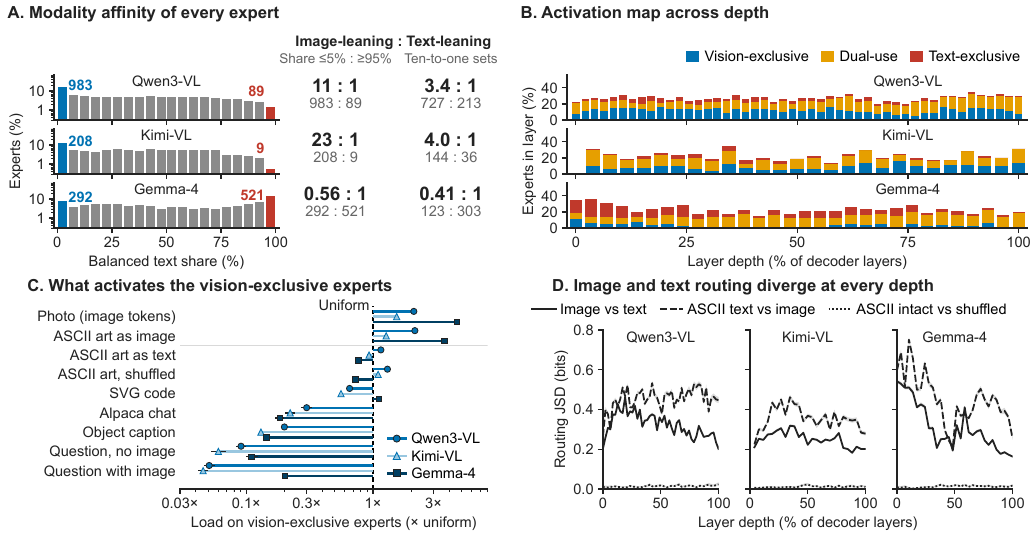}
\caption{\textbf{Routing specialization by modality is real and large, and its direction is model-specific.}
(A) Balanced text share of every routed expert in \qwen, \kimi and \gemma on a log axis, with the counts at the two ends. The table gives the ratio of image-leaning to text-leaning experts under the balanced-share definition (at most 5\% versus at least 95\%) and under the ten-to-one definition of \cref{sec:setup}. The ratio is comparable across the three expert counts, and the two definitions agree on the direction, which \gemma reverses.
(B) Share of each layer's experts that are \vex, dual-use or text-exclusive, over normalized depth.
(C) Load on the \vex experts, as a multiple of the uniform share, for each input type. Image tokens sit at 1.3 to 4.6 times uniform whether the image is a photo or rendered ASCII art, text inside a multimodal prompt at 0.05 to 0.2 times.
(D) Jensen-Shannon divergence between image and text routing per layer, with ASCII text versus image and intact versus shuffled ASCII.}
\label{fig:activation}
\end{figure*}

\paragraph{Lesion and scoped lesion.}
An expert set is removed in one of two modes. Re-route masks the routing scores of the removed experts to $-\infty$ before the top-$k$ selection, so each token goes to its next-best experts. Zero keeps the routing and drops the contribution of removed experts. A scoped lesion applies the mask only at a chosen set of positions: with $M \in \{0,1\}^{L \times E}$ the expert mask and $t(p)$ the modality of position $p$, the router of layer $l$ sees expert $e$ masked at position $p$ only if $M_{le} = 1$ and $t(p)$ is in the scope, image tokens or text tokens, delimiters and generated tokens counting as text. Every lesion is compared with the intact model, with 95\% bootstrap intervals over items, and the \vex lesions with random expert sets of the same size per layer. Changes in accuracy are reported in percentage points, written points.

\section{The Activation Map: Routing Specialization Is Large and Asymmetric}
\label{sec:activation}

Routing separates the modalities sharply (\cref{fig:activation}). In \qwen, 983 of 6,144 routed experts receive at most 5\% of their balanced selections from text tokens, while only 89 receive at least 95\%. \kimi shows the same shape with 208 and 9 of 1,664. \gemma reverses the direction, with 292 against 521, and routes images so narrowly that 16 experts per layer carry 73\% of image selections, against 34\% in \qwen. Under the ten-to-one definition the \vex sets hold 727 (11.8\%), 144 (8.7\%) and 123 (3.2\%) experts. Image tokens put 24.8\% of their selections on the \qwen set, 2.1 times the uniform share of 11.8\%, and text tokens in the same prompts put 0.6\% there, 0.05 times uniform. The image-versus-text routing divergence is 0.2 to 0.5 bits in every layer of every model.

\paragraph{Format control.}
ASCII art given as text reaches the \vex experts at about the uniform rate, well above captions of the same concepts, and shuffling its characters, which destroys recognition, leaves the routing unchanged (\cref{fig:activation}C and D). The experts respond to non-linguistic character runs, not to depicted shapes, and removing them costs ASCIIEval and SVG recognition no more than a random set does (\cref{tab:format}).

\section{Scoped Lesions: Dependence on Experts Sits at Text Positions}
\label{sec:lesion}

\begin{figure*}[t]
\centering
\includegraphics[width=\textwidth]{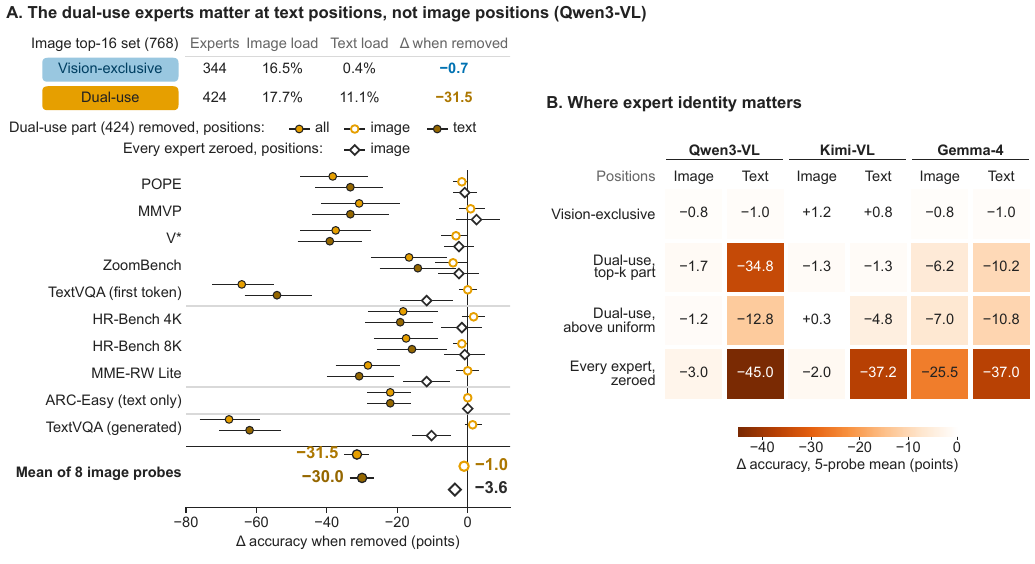}
\caption{\textbf{Experts matter at text positions, not at the image positions that define the modality map.}
(A) Top: the two parts of \qwen's image top-16 set and the cost of removing each. Below: the 424 dual-use experts removed at all positions, at image positions only and at text positions only, over nine probes and greedy TextVQA, with paired 95\% intervals, next to every expert silenced at image positions. Over the eight image probes they cost 31.5, 1.0, 30.0 and 3.6 points: the dual-use cost is at text positions, and every expert at image positions can be silenced for a few points, except for reading text in images.
(B) Five-probe mean cost of removing an expert set at image positions or at text positions, for the three models. Image-side dependence is 3, 2 and 26 points, while text-side dependence on the whole expert population is 37 to 45 points in every model.}
\label{fig:lesion}
\end{figure*}

\begin{figure*}[t]
\centering
\includegraphics[width=\textwidth]{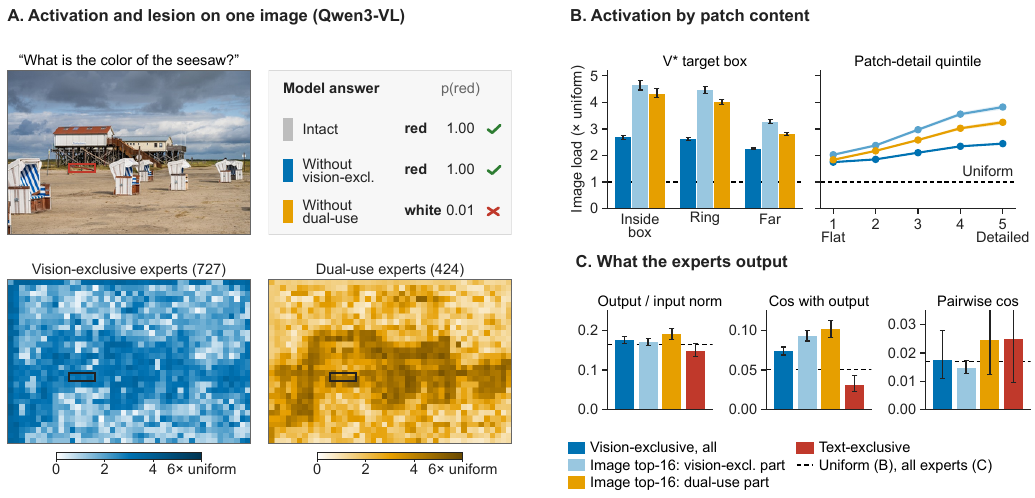}
\caption{\textbf{The \vex experts take informative patches, produce ordinary outputs, and are still dispensable.}
(A) One \vstar item. Load of each patch on the \vex experts and on the dual-use part of the image top-16 set: the share of its expert selections that go to the set, as a multiple of the share under uniform routing, averaged over layers. Intact, the model answers the question correctly with probability 1.00. Without the 727 \vex experts it still does. Without the 424 dual-use experts it answers wrongly.
(B) Load by patch position relative to the \vstar target box, and by patch-detail quintile over all 600 images.
(C) Expert output statistics on image tokens.}
\label{fig:what}
\end{figure*}

\begin{table}[t]
\centering
\footnotesize
\setlength{\tabcolsep}{2.5pt}
\caption{\textbf{Lesions on \qwen.} Accuracy (\%) with each set removed by re-routing unless marked zeroed. Zoom is ZoomBench, TVQA is greedy TextVQA soft accuracy, NLL is caption NLL in nats per token. Random is the mean of five size-matched draws, one draw for TVQA and NLL. The last three rows restrict the lesion to image or text positions.}
\label{tab:lesion}
\fittable{\begin{tabular}{@{}lcccccc@{}}
\toprule
Removed set & POPE & MMVP & \vstar & Zoom & TVQA & NLL \\
\midrule
none & 92.5 & 80.0 & 83.3 & 55.0 & 83.3 & 2.34 \\
\vex (727) & 92.5 & 77.5 & 84.2 & 53.3 & 83.9 & 2.36 \\
random (727) & 92.2 & 79.3 & 77.8 & 50.2 & 83.6 & 2.45 \\
top-16, exclusive part (344) & 92.5 & 79.2 & 82.5 & 49.2 & 86.9 & 2.33 \\
top-16, dual-use part (424) & 54.2 & 49.2 & 45.8 & 38.3 & 15.6 & 3.64 \\
above uniform (737) & 85.8 & 75.8 & 56.7 & 44.2 & 82.2 & 3.37 \\
text top-16 (768) & 90.8 & 56.7 & 43.3 & 35.8 & 50.0 & 3.47 \\
\midrule
dual-use part, image & 90.8 & 80.8 & 80.0 & 50.8 & 84.7 & 2.38 \\
dual-use part, text & 59.2 & 46.7 & 44.2 & 40.8 & 21.4 & 3.45 \\
every expert, image, zeroed & 91.7 & 82.5 & 80.8 & 52.5 & 73.1 & 2.48 \\
\bottomrule
\end{tabular}}
\end{table}

\paragraph{The \vex experts are free.}
Removing every \vex expert of \qwen changes nothing (\cref{tab:lesion}). With all 727 experts re-routed, POPE stays at 92.5\%, \vstar moves from 83.3 to 84.2\%, greedy TextVQA from 83.3 to 83.9\%, and caption likelihood by 0.011 nats per token. Zeroing instead of re-routing gives the same picture, and the three harder probes move by at most 3 points. Five random sets of the same size per layer cost 2.4 points on average over the five main probes and 5 to 6 points on \vstar and ZoomBench, so the \vex experts are easier to remove than arbitrary ones. \kimi (144 experts) and \gemma (123) replicate this, with five-probe mean changes of $+1.8$ and $-1.0$ points against $+0.2$ and $-1.0$ points for random controls. On \gemma only MMVP has an interval that excludes zero, at $-5.8$ points (\cref{tab:gemma}).

\paragraph{The dual-use lesion costs, but only at text positions.}
The 16 experts per layer with the highest image load in \qwen carry 34\% of image-token selections. Their 424 members that text also uses have the same image load as the 344 that only image tokens use (17.7\% against 16.5\%). Removing the 344 costs 0.7 points over eight probes. Removing the 424 costs 31.5 points, 68 points of greedy TextVQA and 1.3 nats of caption likelihood (\cref{fig:lesion}A). Restricting the lesion to one token type locates this cost (\cref{tab:scope_qwen}). Masking the 424 experts only at image-token positions costs 1.0 points, nothing on greedy TextVQA and 0.04 nats, with every per-probe interval including zero. Masking them only at text positions, the question and the answer tokens, reproduces the deficit: 30.0 points, 62 points of greedy TextVQA and 1.11 nats. The same lesion costs 22 points on ARC-Easy, a text-only task. These experts are needed where the answer is computed, not where the image is encoded.

\paragraph{At image positions, expert identity is degenerate.}
Every set we tried can be swapped out at image positions. Re-routing the image top-16, a third of all image routing, costs 0.3 points, and the image top-32, half of it, 0.7. Zeroing the dual-use part at image positions costs 1.2, zeroing the 737 experts that both modalities use above uniform 0.5. Silencing every routed expert at every image position, which removes the whole MoE branch there and leaves the residual stream to attention, costs 3.0 points on the five main probes, 0.14 nats of caption likelihood, nothing on ARC-Easy, and 10 to 12 points only on the two probes that read text in images, TextVQA and MME-RealWorld-Lite. The same silencing at text positions costs 45 points. In \qwen the routed experts do almost nothing at image positions that behavior depends on, and the modality map, which is drawn at those positions, describes computation the model can do without.

\paragraph{\kimi and \gemma.}
\kimi gives the same picture (\cref{tab:scope_kg}, \cref{fig:s1}). The dual-use part costs 1.3 points at either scope, the \vex set nothing, and silencing every routed expert at image positions costs 2.0 points (17 points on TextVQA), against 37.2 points at text positions. \gemma is the model whose image-position experts are functional. Silencing them all costs 25.5 points (55 points on TextVQA), against 37.0 points at text positions, and the dual-use part costs 6.2 points at image positions and 10.2 points at text positions. Its \vex set is still free at both scopes ($-0.8$ and $-1.0$ points). The cost of silencing text positions is 45, 37 and 37 points (\cref{fig:lesion}B), so the three models differ at image positions, not at text positions. Reading text in images is the one ability that needs image-position expert computation in every model.

\begin{figure*}[t]
\centering
\includegraphics[width=\textwidth]{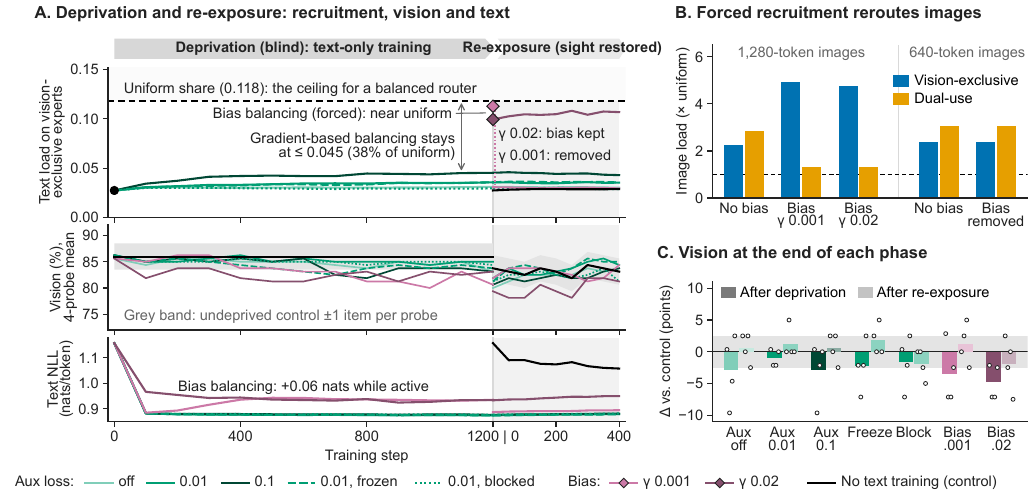}
\caption{\textbf{Text-only training neither recruits the \vex experts nor damages vision through them.}
(A) Text load on the \vex experts, four-probe vision accuracy, and held-out text NLL over 1,200 steps of deprivation (text-only training on UltraChat) and 400 steps of re-exposure (caption training on COCO), for seven balancing conditions and the undeprived control. Diamonds give the load of the bias conditions re-measured with the bias active. In re-exposure the bias is removed at update rate 0.001 and kept at 0.02.
(B) Image load on the \vex and dual-use experts under bias balancing, with the bias active and removed, at matched image budgets.
(C) Vision accuracy relative to the control after deprivation and after re-exposure.}
\label{fig:deprivation}
\end{figure*}

\paragraph{What the \vex experts receive and produce.}
\label{sec:what}
An expert could be dispensable because it sees uninformative input or outputs nothing. Neither holds. Mapping each of the 427k image tokens of the multimodal set back to its patch (\cref{fig:what}), the \vex experts receive the patches inside the \vstar target box at 2.7 times uniform and the background at 2.3 times, and their load rises with patch detail, in step with the dual-use experts. Their outputs are ordinary too. On the MoE inputs of 2,560 image tokens, the \vex experts produce output-to-input norm ratios of 0.175 against 0.191 for the dual-use part and 0.164 over all experts, align with the token's realized MoE output at a cosine of 0.074 against 0.101 and 0.050, and are no more similar to one another than arbitrary experts are. They process informative content with a full-sized update that the rest of the network does not depend on, which is what the image-position lesions above say of every routed expert in this model.

\section{Balancing Is the Pressure That\texorpdfstring{\\}{ }Manufactures Exclusivity}
\label{sec:balancing}

\subsection{Deprivation: The Blind MoE Does Not Learn Braille}
\label{sec:deprivation}

Ordinary text puts 1 to 4\% of its selections on the \vex experts, where the uniform share is 11.8\%. Visual cortex deprived of its input is taken over by the remaining senses \cite{sadato,bavelier}, and the takeover has a cost when sight returns \cite{wieselhubel}. We ask whether a router does the same, and whether a takeover costs anything when the images return.

\paragraph{Design.}
We continue training \qwen on text only, 1,200 steps of UltraChat dialogues \cite{ultrachat} at 16k tokens per step, with rank-8 LoRA \cite{lora} on every expert and trainable routers. Seven conditions differ only in how routing is balanced: the Switch auxiliary loss at coefficients 0, 0.01 and 0.1 \cite{switch}, two variants at 0.01 that freeze the \vex experts or block routing to them during training, and DeepSeek-style bias balancing \cite{auxfree} at update rates 0.001 and 0.02. Every 100 steps we measure held-out text NLL, the routing loads, and POPE, MMVP, \vstar and TextVQA on 40 items each. Each trained state is then re-aligned for 400 steps on COCO captions, next to an undeprived control.

\paragraph{Recruitment is a router effect, and small.}
Text adapts within 100 steps (held-out NLL 1.157 to 0.882) and the vision probes stay within probe noise throughout. The auxiliary loss moves text onto the \vex experts from 2.7\% of selections to 3.1, 3.6 and 4.5\% at coefficients 0, 0.01 and 0.1, against a uniform share of 11.8\%. The shift is identical with the \vex experts frozen and absent when routing to them is blocked: it is the router unmasking experts, not the experts adapting, and text quality is the same in all five conditions. Bias balancing fills the \vex experts to the uniform share (11.3\% of text selections) and redirects image tokens into them, from 27 to 59\% of image selections, at a cost of 0.06 nats of text NLL while the bias is active and at most three probe items on vision.
 After re-exposure, every deprived model is within two items of the undeprived control on all four probes (\cref{fig:deprivation}C, \cref{tab:deprivation}). The one exception keeps its bias through re-exposure (update rate 0.02) and is three items lower on TextVQA. This is what the lesion predicts: capacity that vision did not need cannot be lost to text, and filling it does not help text at this scale.

\subsection{The Balancing Bias of \kimi}
\label{sec:bias}

\begin{figure}[t]
\centering
\includegraphics[width=\linewidth]{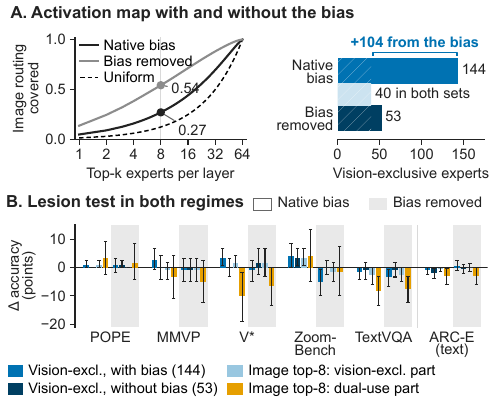}
\caption{\textbf{\kimi's native balancing bias creates most of its \vex experts and changes no behavior.}
(A) Fraction of image routing covered by the top-$k$ experts per layer with and without the native bias, and the number of \vex experts in each regime.
(B) Lesions in both regimes.}
\label{fig:bias}
\end{figure}

\begin{table}[t]
\centering
\footnotesize
\setlength{\tabcolsep}{4pt}
\caption{\textbf{Routing spread, exclusivity and image-side dependence.} $N_{\mathrm{eff}}$: effective experts per layer for image tokens (exponential of the routing entropy). Top-16: share of image routing carried by the 16 highest-load experts per layer. Excl.: share of experts that are \vex. Lesion: five-probe change (points) when they are removed. Silenced: change when every expert is silenced at image positions.}
\label{tab:spread}
\begin{tabular}{lrrrrr}
\toprule
Regime & $N_{\mathrm{eff}}$ & Top-16 & Excl. & Lesion & Silenced \\
\midrule
Qwen3-VL & 96 & 34\% & 11.8\% & $-0.8$ & $-3.0$ \\
Kimi-VL & 55 & 44\% & 8.7\% & $+1.8$ & $-2.0$ \\
Kimi-VL, no bias & 31 & 71\% & 3.2\% & $+0.2$ & $-3.8$ \\
Gemma-4 & 35 & 73\% & 3.2\% & $-1.0$ & $-25.5$ \\
\bottomrule
\end{tabular}
\end{table}

\kimi adds a learned per-expert bias to its routing scores at inference, the aux-loss-free balancing of DeepSeek-V3 \cite{auxfree}. Scaling it to zero asks whether balancing itself produces the exclusive experts (\cref{fig:bias}). Without the bias, the eight experts per layer that image tokens use most cover 54\% of image routing instead of 27\%, and the \vex experts fall from 144 to 53, 40 of them in the native set.

The behavior is unchanged. With the bias removed, every probe moves by at most 5 points, within the noise of 120 items. Lesioning the \vex experts, whether defined with the bias (144) or without it (53), costs at most 5 points on any probe in either regime, and the dual-use members of the image top-8 set cost 5 to 10 points on \vstar and TextVQA in both. This makes the exclusive set a product of load balancing rather than a module that balancing reveals. The same relation holds across models (\cref{tab:spread}): the more balanced the routing, the more experts are exclusive, and removing them costs at most 2 points in every regime. Spread alone does not set image-side dependence either. Without its bias, \kimi routes image tokens as narrowly as \gemma and has the same 3.2\% of exclusive experts, yet silencing every expert at image positions costs 3.8 points against \gemma's 25.5.

\section{Discussion and Conclusion}
\label{sec:discussion}

\paragraph{The modality map is drawn where experts matter least.}
The routing record is accurate: image tokens do prefer a large, stable set of experts that text almost never visits. The lesions show that the dependence on expert identity sits at text positions, where routing is modality-general, and that at image positions, where the modality map is drawn, expert identity is degenerate in the sense of Price and Friston \cite{pricefriston}: with 96 effective experts per layer in \qwen any substitute serves, and even no expert at all costs little.

\paragraph{Consequences for expert selection, pruning and skipping.}
Expert selection for fine-tuning \cite{expertlens}, routing steering \cite{routingdistraction}, and expert pruning and skipping \cite{notallexperts,modes} all operate on the map. Dropping the 727 \vex experts of \qwen removes 12\% of the expert parameters at the same cost as dropping the 727 least-used experts, which is none, and removing both sets, a quarter of all experts, costs at most 5 points on any probe while a random quarter costs 12 points on \vstar (\cref{tab:prune}). And the image-position results explain why modality-aware expert skipping can skip most experts for vision tokens with little loss \cite{modes}: there is little to lose there in \qwen and \kimi, except on OCR, and more in \gemma.

\paragraph{Balancing is the metabolic pressure of MoEs.}
Han et al. \cite{modular} argue that modularity in dense LLMs emerges without the metabolic pressure thought to shape sparse coding in cortex. MoEs have an explicit one: load balancing penalizes any expert that receives less than its share of tokens. Image tokens outnumber text tokens, so the cheapest way to satisfy it is to spread them over experts that text does not need. The pressure produces experts that are busy with images and not depended upon, and it spreads image routing so widely that any expert can stand in for any other at image positions. Degeneracy in the brain is a product of selection for robustness \cite{edelman}, in a MoE of the balancing objective.

\paragraph{Conclusion.}
The experts that a multimodal MoE routes its image tokens to are not the experts its visual behavior depends on. In all three models the \vex experts are dispensable, even in \gemma, where silencing the experts at image positions costs 26 points. The dependence that all three share sits at text positions, in experts that text-only tasks need as well. Load balancing draws the modality map where the models depend on their experts least, and expert selection, pruning and skipping should follow the lesion, not the map. Why \gemma needs its image-position experts when a \kimi with equally narrow routing does not is the question these results leave open.

\section*{LLM Usage Statement}

Claude (Anthropic) was used as an assistant, under the direction of the
authors, for code, the analysis of results, figure scripts, literature
search, discussion of ideas and the text. Claude and ChatGPT (OpenAI) gave reviewer-style feedback on
drafts. The vision-language models analysed in our experiments are
objects of study and are described in the main text. The authors
verified all results and take full responsibility for the content.

{
    \small
    \bibliographystyle{ieeenat_fullname}
    \bibliography{main}
}
\clearpage
\maketitlesupplementary
\setcounter{figure}{0}
\renewcommand{\thefigure}{S\arabic{figure}}
\setcounter{table}{0}
\renewcommand{\thetable}{S\arabic{table}}
\setcounter{section}{0}
\renewcommand{\thesection}{\Alph{section}}
\renewcommand{\theHsection}{S\Alph{section}}
\raggedbottom

This document gives the full results behind the main paper and the settings of its training runs.

\section{Format control}
\label{sec:suppl_format}

If the \vex experts implemented visual computation in a format-independent way, text that depicts shapes should reach them. ASCII art presented as text routes to the \vex set at 0.8 to 1.2 times uniform in the three models, well above captions of the same concepts, but shuffling the characters of each ASCII line destroys recognition and leaves routing untouched (Fig.~\ref*{fig:activation}C and D of the main paper). The Jensen-Shannon divergence is 0.011 to 0.014 bits between intact and shuffled ASCII and 0.3 to 0.5 bits to any other text group.

\cref{tab:format} gives the recognition accuracy on the format probes of Sec.~\ref*{sec:activation} of the main paper for \qwenfull, intact and under four lesions: the \vex experts removed by re-routing and by zeroing, random sets of the same size, and the text-exclusive experts removed by re-routing. Item counts are in parentheses. Photos exist for 27 of the 300 ASCIIEval concepts, and each synthetic shape is given as a raster image, as SVG code, as ASCII text, as rendered ASCII (as image) and as a caption.

\begin{table}[h]
\centering
\footnotesize
\setlength{\tabcolsep}{2.5pt}
\caption{\textbf{Format probes on \qwen.} Four-way recognition accuracy (\%) under each lesion. Random is the mean of two size-matched draws.}
\label{tab:format}
\fittable{\begin{tabular}{@{}lccccc@{}}
\toprule
 & & \multicolumn{2}{c}{\vex} & & text-excl. \\
\cmidrule(lr){3-4}
Probe & none & reroute & zero & random & reroute \\
\midrule
ASCIIEval photo (27) & 92.6 & 92.6 & 92.6 & 92.6 & 96.3 \\
ASCIIEval as image (300) & 60.7 & 63.3 & 61.7 & 61.0 & 61.0 \\
ASCIIEval as text (300) & 31.7 & 30.3 & 29.3 & 29.3 & 32.0 \\
ASCIIEval caption (300) & 99.7 & 99.7 & 99.7 & 100.0 & 100.0 \\
shapes raster (48) & 100.0 & 100.0 & 100.0 & 100.0 & 100.0 \\
shapes as image (48) & 91.7 & 91.7 & 89.6 & 86.5 & 87.5 \\
shapes as text (48) & 56.2 & 47.9 & 43.8 & 49.0 & 50.0 \\
shapes SVG (48) & 87.5 & 85.4 & 85.4 & 79.2 & 79.2 \\
shapes caption (48) & 100.0 & 100.0 & 100.0 & 100.0 & 100.0 \\
\bottomrule
\end{tabular}}
\end{table}

Removing the 213 text-exclusive experts cuts caption log-odds from 13.6 to 10.2 and SVG log-odds by 2.7, so the lesion machinery detects text-side damage where it exists.

\section{Deprivation conditions and training settings}
\label{sec:suppl_deprivation}

\cref{tab:deprivation} lists the end state of the seven conditions of Sec.~\ref*{sec:deprivation} of the main paper. Aux off, aux 0.01 and aux 0.1 train with the Switch auxiliary loss at that coefficient. Freeze and block use coefficient 0.01, and freeze the adapters of the \vex experts or mask their router logits during training. Bias 0.001 and bias 0.02 use bias balancing at that update rate without the auxiliary loss, and their load is measured with the bias active.

\begin{table}[h]
\centering
\footnotesize
\setlength{\tabcolsep}{3pt}
\caption{\textbf{Deprivation and re-exposure on \qwen.} Load is the text load on the \vex experts after 1,200 text-only steps (uniform 0.118), NLL the held-out text NLL, and POPE / MMVP / \vstar / TextVQA (40 items each) after deprivation and after 400 caption steps. The undeprived control ends re-exposure at 85.0 / 80.0 / 77.5 / 90.0.}
\label{tab:deprivation}
\fittable{\begin{tabular}{@{}lcccc@{}}
\toprule
Condition & Load & NLL & After deprivation & After re-exposure \\
\midrule
aux off & 0.031 & 0.874 & 85.0/75.0/85.0/87.5 & 87.5/80.0/80.0/87.5 \\
aux 0.01 & 0.036 & 0.874 & 85.0/82.5/87.5/85.0 & 85.0/85.0/77.5/90.0 \\
aux 0.1 & 0.045 & 0.876 & 85.0/75.0/87.5/85.0 & 87.5/80.0/80.0/87.5 \\
freeze & 0.036 & 0.874 & 82.5/77.5/87.5/87.5 & 87.5/80.0/82.5/90.0 \\
block & 0.029 & 0.876 & 85.0/77.5/87.5/87.5 & 85.0/80.0/75.0/85.0 \\
bias 0.001 & 0.113 & 0.936 & 87.5/77.5/82.5/82.5 & 85.0/82.5/82.5/87.5 \\
bias 0.02 & 0.099 & 0.934 & 82.5/77.5/82.5/82.5 & 87.5/77.5/77.5/82.5 \\
\bottomrule
\end{tabular}}
\end{table}

\paragraph{Deprivation.}
Every routed expert of \qwen carries a LoRA adapter of rank 8 ($\alpha = 16$), and the router weights are trained directly. The loss is the next-token loss on the assistant turns of UltraChat dialogues, with 8 dialogues of up to 2,048 tokens per step. We use AdamW without weight decay, learning rates of $10^{-4}$ for the adapters and $5 \times 10^{-5}$ for the routers, a linear warm-up over 50 steps and gradient clipping at norm 1.

\paragraph{Re-exposure.}
The same parameters are trained for 400 steps on COCO captions, with the loss on the caption tokens, 8 image-caption pairs per step, images at up to 640 visual tokens and a warm-up over 20 steps. The images of the vision probes are excluded.

\paragraph{Measurements.}
Held-out text NLL uses 40 UltraChat dialogues from the test split. Routing loads are measured on 30 Alpaca prompts and on 20 image-question pairs. Each vision probe has 40 items, so one item is 2.5 points.

\twocolumn[{%
{\centering\footnotesize\setlength{\tabcolsep}{3pt}%
\captionof{table}{\textbf{Pruning by routing affinity on \qwen.} The \vex set and the least-used set each have 727 experts, matched per layer, and are disjoint. Their union holds 1,454 experts (24\%). The affinity-462 set is the 462 experts with the lowest raw text share, the vision-expert count reported for this model by ExpertLens (Marsili et al., 2026). Accuracy (\%), caption NLL in nats per token, greedy TextVQA soft accuracy (\%). Random sets are size-matched per layer (mean of two draws for the 727 case, one for the generation metrics and for the 1,454 case).}
\label{tab:prune}
\vspace{2pt}
\fittable{\begin{tabular}{lccccccc}
\toprule
Probe & none & \vex (727) & least-used (727) & random (727) & union (1,454) & random (1,454) & affinity-462 \\
\midrule
POPE & 92.5 & 92.5 & 92.5 & 92.5 & 92.5 & 93.3 & 92.5 \\
MMVP & 80.0 & 77.5 & 77.5 & 81.3 & 77.5 & 79.2 & 79.2 \\
\vstar & 83.3 & 84.2 & 83.3 & 79.6 & 83.3 & 71.7 & 83.3 \\
ZoomBench & 55.0 & 53.3 & 51.7 & 47.5 & 50.0 & 47.5 & 55.8 \\
TextVQA (first token) & 84.2 & 84.2 & 84.2 & 83.8 & 82.5 & 84.2 & 83.3 \\
HR-Bench 4K & 65.8 & 68.3 & 70.8 & 66.7 & 67.5 & 61.7 & 66.7 \\
HR-Bench 8K & 58.3 & 60.0 & 59.2 & 59.2 & 60.0 & 60.0 & 58.3 \\
MME-RealWorld-Lite & 54.2 & 52.5 & 50.0 & 54.2 & 51.7 & 48.3 & 55.8 \\
ARC-Easy (text) & 98.5 & 98.0 & 97.5 & 97.5 & 97.5 & 96.5 & 98.5 \\
Caption NLL & 2.34 & 2.36 & 2.35 & 2.45 & 2.36 & 2.29 & 2.33 \\
TextVQA (generated) & 83.3 & 83.9 & 82.2 & 83.6 & 83.1 & 84.4 & 83.6 \\
\bottomrule
\end{tabular}}\par}
\vspace{1.4em}
{\centering\footnotesize\setlength{\tabcolsep}{3pt}%
\captionof{table}{\textbf{Lesions on \gemma.} Accuracy (\%) with each set removed by re-routing. The image top-16 holds 480 experts, of which 84 are \vex (image load 13\%) and 396 dual-use (60\%). Caption NLL in nats per token.}
\label{tab:gemma}
\vspace{2pt}
\fittable{\begin{tabular}{lcccccc}
\toprule
Probe & none & \vex (123) & random (123) & least-used (123) & top-16, \vex part & top-16, dual-use part \\
\midrule
POPE & 90.8 & 90.8 & 90.0 & 90.8 & 90.0 & 77.5 \\
MMVP & 87.5 & 81.7 & 85.0 & 86.7 & 81.7 & 63.3 \\
\vstar & 60.0 & 54.2 & 59.2 & 64.2 & 52.5 & 44.2 \\
ZoomBench & 42.5 & 46.7 & 44.2 & 43.3 & 47.5 & 36.7 \\
TextVQA (first token) & 76.7 & 79.2 & 74.2 & 77.5 & 79.2 & 68.3 \\
HR-Bench 4K & 64.2 & 67.5 & 65.8 & 63.3 & 65.8 & 45.8 \\
HR-Bench 8K & 52.5 & 47.5 & 50.8 & 50.8 & 50.0 & 37.5 \\
MME-RealWorld-Lite & 41.7 & 40.0 & 44.2 & 44.2 & 39.2 & 24.2 \\
ARC-Easy (text) & 98.0 & 97.5 & 97.0 & 98.0 & 97.5 & 92.5 \\
Caption NLL & 4.29 & 4.08 & 4.31 & 4.27 & 4.07 & 5.04 \\
TextVQA (generated) & 70.8 & 73.3 & 70.0 & 71.7 & 71.7 & 63.6 \\
\bottomrule
\end{tabular}}\par}
\vspace{1.4em}
}]

\section{Pruning panel}
\label{sec:suppl_prune}

\cref{tab:prune} compares ways of choosing which experts of \qwen to remove, at matched size per layer. The \vex set and the 727 least-used experts are disjoint, and their union removes 24\% of the pool. Under each of the three, no probe loses more than 5 points and caption NLL moves by at most 0.02 nats. A random set of 1,454 experts loses 11.6 points on \vstar and 7.5 on ZoomBench. The affinity-462 set changes no probe by more than 2 points.

\newpage
\section{Lesions on \gemma}
\label{sec:suppl_gemma}

\cref{tab:gemma} gives the lesions at all positions on \gemmafull. Removing its 123 \vex experts changes the five-probe mean by $-1.0$ points, the same as a random set of 123 experts. The dual-use part of the image top-16 set costs 13.5 points.

\twocolumn[{%
{\centering%
\includegraphics[width=\textwidth]{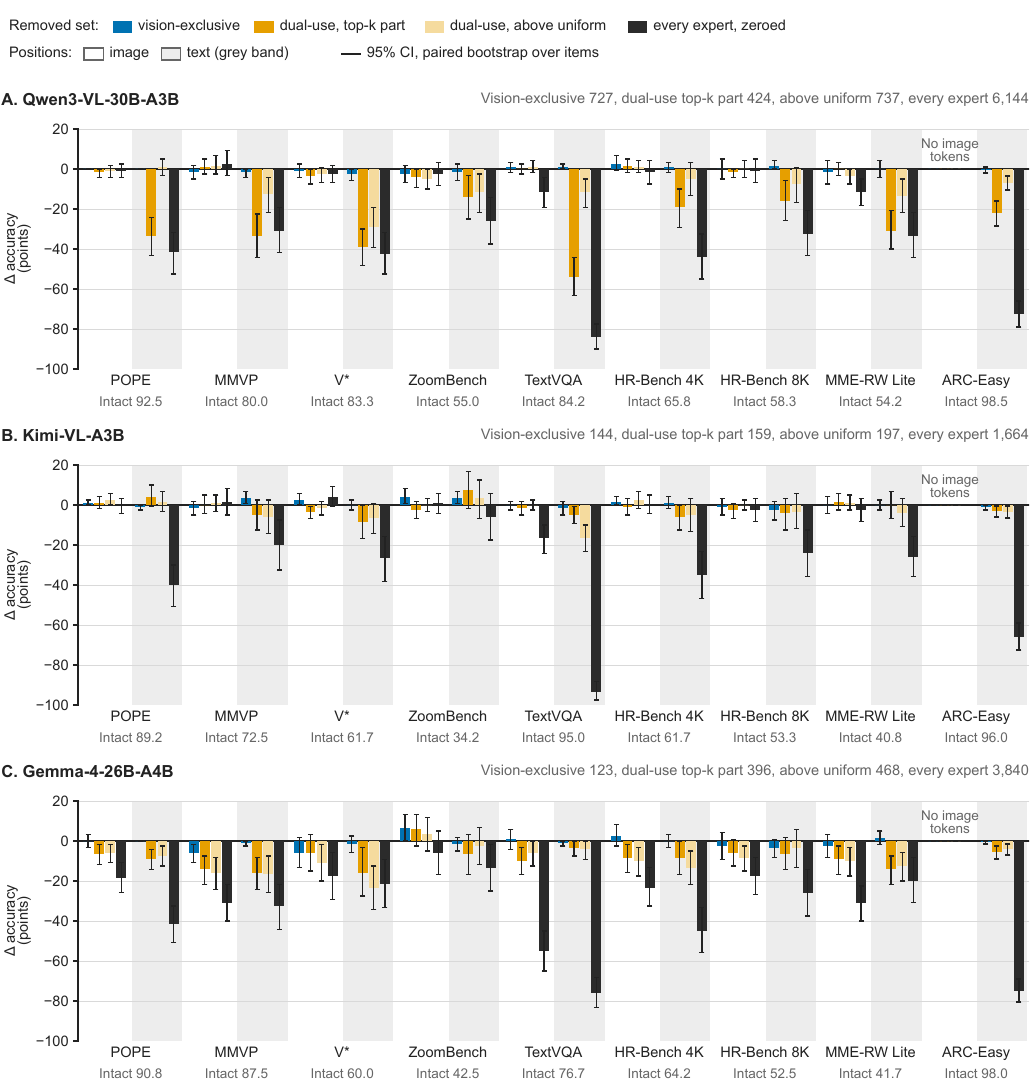}
\captionof{figure}{\textbf{Token-scoped lesions of four expert sets on every probe.} Change in accuracy against the intact model when the set is removed only at image-token positions (white background) or only at text positions (grey band), for the three models, with paired 95\% bootstrap intervals. The intact accuracy and the set sizes are printed per panel. ARC-Easy has no image tokens.}
\label{fig:s1}
\par}
\vspace{1.4em}
}]

\section{Token-scoped lesions}
\label{sec:suppl_scope}

\cref{fig:s1} plots four expert sets of \cref{tab:scope_qwen,tab:scope_kg} under the native routing. The lesions on \kimifull without its balancing bias are in \cref{tab:scope_kg} only.

Greedy TextVQA on \gemma: $-7.2$ points for the dual-use part at all positions, $-6.4$ at image positions, $-1.9$ at text positions, $-52.2$ with every expert silenced at image positions. Caption NLL changes by $+0.75$, $+0.04$, $+0.47$ and $+0.28$ nats per token.

\twocolumn[{%
{\centering\footnotesize\setlength{\tabcolsep}{2.5pt}%
\captionof{table}{\textbf{Token-scoped lesions on \qwen.} Change in accuracy (points) against the intact model with the set masked at all positions, only at image-token positions or only at text positions (prompt and generated tokens), re-routed unless marked zeroed. Caption NLL change in nats per token, greedy TextVQA change in points.}
\label{tab:scope_qwen}
\vspace{2pt}
\fittable{\begin{tabular}{lcccccccccccc}
\toprule
Lesion & POPE & MMVP & \vstar & Zoom & TextVQA & HR-4K & HR-8K & MME-RW & ARC & Mean 5 & Cap. NLL & TextVQA gen. \\
\midrule
dual-use part (424), all & $-38.3$ & $-30.8$ & $-37.5$ & $-16.7$ & $-64.2$ & $-18.3$ & $-17.5$ & $-28.3$ & $-22.0$ & $-37.5$ & $+1.30$ & $-67.8$ \\
dual-use part, image & $-1.7$ & $+0.8$ & $-3.3$ & $-4.2$ & $0.0$ & $+1.7$ & $-1.7$ & $0.0$ & $0.0$ & $-1.7$ & $+0.04$ & $+1.4$ \\
dual-use part, text & $-33.3$ & $-33.3$ & $-39.2$ & $-14.2$ & $-54.2$ & $-19.2$ & $-15.8$ & $-30.8$ & $-22.0$ & $-34.8$ & $+1.11$ & $-61.9$ \\
dual-use part, image, zeroed & $0.0$ & $0.0$ & $-2.5$ & $-3.3$ & $0.0$ & $+4.2$ & $-2.5$ & $-2.5$ & $0.0$ & $-1.2$ & $+0.05$ & $+1.7$ \\
\vex (727), image & $0.0$ & $-1.7$ & $-0.8$ & $-2.5$ & $+0.8$ & $+2.5$ & $0.0$ & $-1.7$ & $0.0$ & $-0.8$ & $0.00$ & $+1.4$ \\
\vex (727), text & $0.0$ & $-1.7$ & $-2.5$ & $-1.7$ & $+0.8$ & $+0.8$ & $+1.7$ & $0.0$ & $-0.5$ & $-1.0$ & $+0.01$ & $+1.7$ \\
above uniform (737), image & $-0.8$ & $+1.7$ & $-2.5$ & $-5.0$ & $+0.8$ & $+0.8$ & $0.0$ & $-3.3$ & $0.0$ & $-1.2$ & $0.00$ & $+0.6$ \\
above uniform (737), text & $+0.8$ & $-12.5$ & $-29.2$ & $-11.7$ & $-11.7$ & $-5.0$ & $-7.5$ & $-13.3$ & $-7.0$ & $-12.8$ & $+0.90$ & $0.0$ \\
image top-16 (768), image & $0.0$ & $-0.8$ & $-1.7$ & $-2.5$ & $+3.3$ & $0.0$ & $0.0$ & $-5.8$ & $0.0$ & $-0.3$ & $+0.03$ & $+3.9$ \\
image top-32 (1,536), image & $-0.8$ & $0.0$ & $0.0$ & $-3.3$ & $+0.8$ & $+1.7$ & $+0.8$ & $-7.5$ & $0.0$ & $-0.7$ & $+0.05$ & $+2.2$ \\
image top-16 (768), text & $-32.5$ & $-29.2$ & $-39.2$ & $-12.5$ & $-64.2$ & $-20.0$ & $-15.8$ & $-31.7$ & $-25.5$ & $-35.5$ & $+1.37$ & $-69.4$ \\
every expert, image, zeroed & $-0.8$ & $+2.5$ & $-2.5$ & $-2.5$ & $-11.7$ & $-1.7$ & $-0.8$ & $-11.7$ & $0.0$ & $-3.0$ & $+0.14$ & $-10.3$ \\
every expert, text, zeroed & $-41.7$ & $-30.8$ & $-42.5$ & $-25.8$ & $-84.2$ & $-44.2$ & $-32.5$ & $-33.3$ & $-72.5$ & $-45.0$ & $+12.13$ & $-83.3$ \\
\bottomrule
\end{tabular}}\par}
\vspace{1.4em}
{\centering\footnotesize\setlength{\tabcolsep}{2.5pt}%
\captionof{table}{\textbf{Token-scoped lesions on \kimi, on \kimi with its balancing bias removed, and on \gemma.} Change in accuracy (points) against the intact model of the same regime. Five-probe mean over POPE, MMVP, \vstar, ZoomBench and TextVQA. In the no-bias block the dual-use part is the set defined under the native routing and the \vex set is the one defined without the bias.}
\label{tab:scope_kg}
\vspace{2pt}
\fittable{\begin{tabular}{llcccccccccc}
\toprule
Model & Lesion & POPE & MMVP & \vstar & Zoom & TextVQA & HR-4K & HR-8K & MME-RW & ARC & Mean 5 \\
\midrule
\kimi & dual-use part (159), all & $+3.3$ & $-3.3$ & $-10.0$ & $+4.2$ & $-8.3$ & $-10.0$ & $-6.7$ & $0.0$ & $-3.0$ & $-2.8$ \\
 & dual-use part, image & $+0.8$ & $0.0$ & $-3.3$ & $-2.5$ & $-1.7$ & $-0.8$ & $-2.5$ & $+1.7$ & $0.0$ & $-1.3$ \\
 & dual-use part, text & $+4.2$ & $-5.0$ & $-8.3$ & $+7.5$ & $-5.0$ & $-5.8$ & $-4.2$ & $0.0$ & $-3.0$ & $-1.3$ \\
 & \vex (144), image & $+0.8$ & $-1.7$ & $+2.5$ & $+4.2$ & $0.0$ & $+1.7$ & $-0.8$ & $0.0$ & $0.0$ & $+1.2$ \\
 & \vex (144), text & $-0.8$ & $+3.3$ & $0.0$ & $+3.3$ & $-1.7$ & $+0.8$ & $-2.5$ & $0.0$ & $-1.0$ & $+0.8$ \\
 & above uniform (197), image & $+2.5$ & $+0.8$ & $-1.7$ & $0.0$ & $0.0$ & $+2.5$ & $0.0$ & $+0.8$ & $0.0$ & $+0.3$ \\
 & above uniform (197), text & $+1.7$ & $-5.8$ & $-6.7$ & $+3.3$ & $-16.7$ & $-5.0$ & $-3.3$ & $-4.2$ & $-3.5$ & $-4.8$ \\
 & every expert, image, zeroed & $0.0$ & $+1.7$ & $+4.2$ & $+0.8$ & $-16.7$ & $0.0$ & $-2.5$ & $-2.5$ & $0.0$ & $-2.0$ \\
 & every expert, text, zeroed & $-40.0$ & $-20.0$ & $-26.7$ & $-5.8$ & $-93.3$ & $-35.0$ & $-24.2$ & $-25.8$ & $-66.0$ & $-37.2$ \\
\midrule
\kimi, no bias & dual-use part (159), image & $+0.8$ & $-0.8$ & $-0.8$ & $-0.8$ & $-3.3$ & $0.0$ & $-0.8$ & $+0.8$ & $0.0$ & $-1.0$ \\
 & dual-use part (159), text & $+3.3$ & $-5.0$ & $-5.0$ & $-2.5$ & $-5.0$ & $-6.7$ & $0.0$ & $-1.7$ & $-3.0$ & $-2.8$ \\
 & \vex (53), image & $0.0$ & $-0.8$ & $+1.7$ & $+0.8$ & $-0.8$ & $+0.8$ & $+0.8$ & $-0.8$ & $0.0$ & $+0.2$ \\
 & \vex (53), text & $+1.7$ & $+0.8$ & $-1.7$ & $+1.7$ & $-1.7$ & $+1.7$ & $+0.8$ & $0.0$ & $-0.5$ & $+0.2$ \\
 & every expert, image, zeroed & $+1.7$ & $-0.8$ & $+1.7$ & $-1.7$ & $-20.0$ & $+0.8$ & $-5.0$ & $-4.2$ & $0.0$ & $-3.8$ \\
\midrule
\gemma & dual-use part (396), all & $-13.3$ & $-24.2$ & $-15.8$ & $-5.8$ & $-8.3$ & $-18.3$ & $-15.0$ & $-17.5$ & $-5.5$ & $-13.5$ \\
 & dual-use part, image & $-6.7$ & $-14.2$ & $-5.8$ & $+5.8$ & $-10.0$ & $-8.3$ & $-5.8$ & $-9.2$ & $0.0$ & $-6.2$ \\
 & dual-use part, text & $-9.2$ & $-15.8$ & $-15.8$ & $-6.7$ & $-3.3$ & $-8.3$ & $-6.7$ & $-14.2$ & $-5.5$ & $-10.2$ \\
 & \vex (123), image & $0.0$ & $-5.8$ & $-5.8$ & $+6.7$ & $+0.8$ & $+2.5$ & $-2.5$ & $-2.5$ & $0.0$ & $-0.8$ \\
 & \vex (123), text & $0.0$ & $-0.8$ & $-1.7$ & $-1.7$ & $-0.8$ & $0.0$ & $-3.3$ & $+1.7$ & $-0.5$ & $-1.0$ \\
 & above uniform (468), image & $-5.8$ & $-15.8$ & $-10.8$ & $+3.3$ & $-5.8$ & $-10.0$ & $-8.3$ & $-10.0$ & $0.0$ & $-7.0$ \\
 & above uniform (468), text & $-7.5$ & $-16.7$ & $-23.3$ & $-2.5$ & $-4.2$ & $-13.3$ & $-3.3$ & $-12.5$ & $-4.0$ & $-10.8$ \\
 & every expert, image, zeroed & $-18.3$ & $-30.8$ & $-17.5$ & $-5.8$ & $-55.0$ & $-23.3$ & $-17.5$ & $-30.8$ & $0.0$ & $-25.5$ \\
 & every expert, text, zeroed & $-41.7$ & $-32.5$ & $-21.7$ & $-13.3$ & $-75.8$ & $-45.0$ & $-25.8$ & $-20.0$ & $-75.0$ & $-37.0$ \\
\bottomrule
\end{tabular}}\par}
\vspace{1.4em}
}]

\section{Controls and intervals}
\label{sec:suppl_stats}

A random control draws, in every layer, as many experts as the lesion removes in that layer, from the experts outside the lesioned set. Intervals are 95\% percentile intervals over 2,000 bootstrap resamples of the items, paired between the lesioned and the intact model. For a mean over probes, the items are resampled within each probe.

The text scope includes the vision delimiter tokens. Lesions restricted to the delimiters change the five-probe mean by at most 1.7 points in every model.

\section{Reproducibility}
\label{sec:suppl_repro}

All routing statistics, lesions and training runs use the public checkpoints of the three models. Code is available at \coderepo. One H200 GPU suffices for every experiment. A routing pass over the 2,767 probe items takes 13 minutes on \qwen, the full lesion suite 2 hours, one deprivation run 3.2 hours, and the complete \gemma pipeline 65 minutes.

\end{document}